\documentclass[journal,twoside]{IEEEtran}
\usepackage[]{graphicx}
\usepackage{psfrag} 
\usepackage{amsmath,amsthm} 
\usepackage[x11names]{xcolor}
\usepackage{multirow}
\usepackage{rotating}
\usepackage{overpic}
\usepackage{hyperref}
\usepackage{enumerate}
\usepackage{array} 
\usepackage{graphicx} 
\usepackage{subfigure}
\usepackage{caption}    
\usepackage{color}

\newcommand{\cb}[1]{\ifmmode {\boldsymbol{#1}}\else ${\boldsymbol{#1}}$\fi}
\newcommand{\cp}[1]{\ifmmode {\mathcal{#1}}\else ${\mathcal{#1}}$\fi}

\newcommand{\bH}{\cb{H}}
\newcommand{\bx}{\cb{x}}
\newcommand{\bX}{\cb{X}}
\newcommand{\bE}{\cb{E}}
\newcommand{\be}{\cb{e}}
\newcommand{\bA}{\cb{A}}
\newcommand{\ba}{\cb{a}}
\newcommand{\bW}{\cb{W}}
\newcommand{\bz}{\cb{z}}

\newcommand{\red}{\color{red}}

\newcommand{\gray}{\color{gray}}
\newtheorem*{proposition}{Proposition}

\theoremstyle{definition}
\newtheorem*{definition}{Definition}

\theoremstyle{definition}

\begin{document}
%
\title{Kernel Nonnegative Matrix Factorization \\Without the Curse of the Pre-image --- \\Application to Unmixing Hyperspectral Images}
%
%
%

\author{Fei Zhu, Paul Honeine, 
Maya Kallas
\thanks{F. Zhu is with the Institut Charles Delaunay (CNRS), Universit\'{e} de Technologie de Troyes, Troyes, France.
e-mail: fei.zhu@utt.fr.}
\thanks{P. Honeine is with the LITIS lab, Universit\'{e} de Rouen, Rouen, France.
e-mail: paul.honeine@univ-rouen.fr.}%
\thanks{M. Kallas is with the Centre de Recherche en Automatique de Nancy (CNRS), Universit\'{e} de Lorraine, Nancy, France.
e-mail: maya.kallas@univ-lorraine.fr}
}

%
%

\markboth{}
{Zhu \MakeLowercase{\textit{et al.}}: Kernel-NMF without the curse of the pre-image}
%



\maketitle

\begin{abstract}
The nonnegative matrix factorization (NMF) is widely used in signal and image processing, including bio-informatics, blind source separation and hyperspectral image analysis in remote sensing. A great challenge arises when dealing with a nonlinear formulation of the NMF. Within the framework of kernel machines, the models suggested in the literature do not allow the representation of the factorization matrices, which is a fallout of the curse of the pre-image. In this paper, we propose a novel kernel-based model for the NMF that does not suffer from the pre-image problem, by investigating the estimation of the factorization matrices directly in the input space. For different kernel functions, we describe two schemes for iterative algorithms: an additive update rule based on a gradient descent scheme and a multiplicative update rule in the same spirit as in the Lee and Seung algorithm. Within the proposed framework, we develop several extensions to incorporate constraints, including sparseness, smoothness, and spatial regularization with a total-variation-like penalty. The effectiveness of the proposed method is demonstrated with the problem of unmixing hyperspectral images, using well-known real images and results with state-of-the-art techniques.
\end{abstract}


%
\IEEEpeerreviewmaketitle

\section{Introduction}
%
%
%
%
\IEEEPARstart{T}{he nonnegative} matrix factorization (NMF) has become a prominent analysis technique in many fields, owing to its power to extract sparse and tractable interpretable representations from a given data matrix. The scope of application spans feature extraction, compression and visualization, within pattern recognition, machine learning, and signal and image processing \cite{handbookBSS,NMF_WhyHow}. It has been popularized since Lee and Seung discovered that, when applied to an image, ``NMF is able to learn the parts of objects'' \cite{lee99}. Since then, NMF has been successfully applied in image classification \cite{Buchsbaum02,Guillamet01}, face expression recognition \cite{LiHou01,Buciu04}, audio 
analysis \cite{Smaragdis04, Lefevre2011}, objet recognition \cite{Liu04,Wild04}, computational biology \cite{Devarajan2008}, gene expression data \cite{Brunet2004,kim03}, and clustering \cite{Young2006}. Moreover, the NMF is tightly connected to spectral clustering \cite{Xu03,Ding2005,Li2006}. See also \cite{cichocki2009nonnegative} for a review. 

The NMF consists in approximating a nonnegative matrix with two low-rank nonnegative ones. It allows a sparse representation with nonnegativity constraints, which often provides a physical interpretation to the factorization thanks to the resulting part-based representation, as opposed to conventional subtractive models. Typically, this idea is described with the issue of spectral unmixing in hyperspectral imagery, as illustrated next. A hyperspectral image details the scene under scrutiny with spectral observations of electromagnetic waves emitted/reflected from it. Typically, it corresponds to the acquisition of a ground scene from which sunlight is reflected. A hyperspectral image consists of a three-dimensional data cube, two of the dimensions being spatial, and the third one being the reflectance. In other words, a spectral characteristic is available at each pixel. For example, the AVIRIS sensors have 224 contiguous spectral bands, covering from 0.4 to 2.5 $\mu$m, with a ground resolution that varies from 4 to 20 m (depending on the distance of the airborne to the ground). Due to such spatial resolution, any acquired spectrum is a superposition of spectra of several underlying materials. The (spectral) unmixing of a given hyperspectral image aims to extract the spectra of these single ``pure'' materials, called endmembers, and to estimate the abundance of each endmember in every pixel, {\em i.e.,} every position of the area under scrutiny. It is obvious that both abundances and spectra of endmembers are nonnegative. The NMF provides a decomposition suitable for such physical interpretation.



The physical interpretation of the NMF is however not for free. To illustrate this, consider the well-known singular-value-decomposition (SVD), which allows to solve efficiently the unconstrained matrix factorization problem with orthogonality constraints, under the risk of losing the physical meaning while providing a unique solution. It is known that the SVD has polynomial-time complexity. As opposed to the SVD, the NMF is unfortunately a NP-hard and an ill-posed problem, in general. In fact, it is proven in \cite{NMFcomplexity} that the NMF is NP-hard
; see also \cite{Gillis2011}. NMF is ill-posed, as illustrated by the fact that the decomposition is not unique; see \cite{NMFuniqueness} and references therein. In practice, the non-uniqueness issue is alleviated by including priors other than the nonnegativity, the most known being sparseness and smoothness constraints.




First studied in the 1977 in \cite{Leggett77}, the NMF problem was reinvented several times, scilicet with the work of Paatero and Tapper in \cite{nmfPaatero}. It has gained popularity thanks to the work of Lee and Seung published in {\em Nature} \cite{lee99}. Many optimization algorithms have been proposed for NMF, such as the multiplicative update rules \cite{Lee2001} and nonnegative least squares \cite{Kim2008}. Sparseness, which allows the uniqueness and enhances interpretation, is often imposed either with projections \cite{Hoyer04} or with $\ell_1$-norm regularization \cite{sparseNMF}. Smoothness also reduces the degrees of freedom, typically in the spectral unmixing problem, either by using piecewise smoothness of the estimated endmembers \cite{Pauca2006,Jia2009,Qian2011}, or by favoring spatial coherence with a regularization similar to the total-variation (TV) penalty \cite{Iordache2012}. Additional constraints are the orthogonality \cite{Ding06,Li2007}, the minimum-volume \cite{Zhou2011}, and the sum-to-one constraint which is often imposed on the abundances \cite{Masalmah2008}. As illustrated in all these developments with the unmixing problem in hyperspectral imagery, the NMF and most of its variants are based on a linear mixing assumption. Providing nonlinear models for NMF is a challenging issue 
\cite{Yang2012}.

Kernel machines have been offering an elegant framework to derive nonlinear techniques based on linear ones, by mapping the data using some nonlinear function to a feature space, and applying the linear algorithm on the mapped data \cite{Shawetaylor_Cristianini}. The key idea is the kernel trick, where a kernel function allows to evaluate the inner product between transformed data without the need of an explicit knowledge of the mapping function. This trick allows to easily extend the mapping to functional spaces, {\em i.e., reproducing kernel Hilbert space}, and infinite dimensional spaces, namely when using the prominent Gaussian kernel. Kernel machines have been widely used for decisional tasks, initially with the so-called support vector machines for classification and regression \cite{Vap95}. Unsupervised learning has been tackled in \cite{KPCA} with the kernel principal component analysis (KPCA), and more recently in \cite{KECA} with the kernel entropy component analysis. It is worth noting that an attractive property of kernel machines is that the use of the linear inner product kernel leads to the underlying conventional linear technique.

Recently, a few attempts have been made to derive a kernel-NMF, for the sake of a nonlinear variant of the conventional NMF \cite{Zhang2006,Ding10,Li2012}. To this end, the linear model in the latter is defined by writing each column of the matrix under scrutiny as the linear combination of the columns of the first matrix to be determined, the second matrix being defined by the weights of the linear combination. By defining the input space with the columns of the studied matrix, these columns are mapped with a nonlinear transformation to some feature space where the linear model is applied. Unfortunately, the obtained results cannot be exploited, since the columns of the first unknown matrix lie in the feature space. One needs to get back from the (often infinite dimensional) feature space to the input space. This is the curse of the pre-image problem, a major drawback inherited from kernel machines \cite{11.spm}. It was first revealed in denoising with KPCA, where the denoised feature should be mapped back to the input space \cite{Mika99kernelpca}. This ill-posed problem yields an even more difficult problem when dealing with the nonnegativity of the result \cite{13.pr.nn_preimage}.

\begin{figure}[t]
\bigskip
\centering
  \graphicspath{{Graphics/}}
  \begin{overpic}[width=0.42\textwidth]{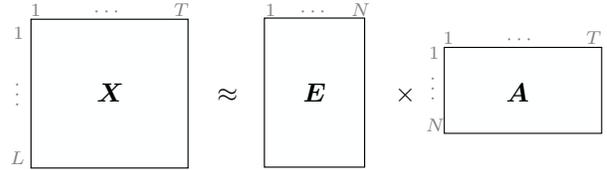}
       \put(-3.5,1){\gray \scriptsize $L$}
       \put(10,12){ $\bX$}
       \put(-1.5,11){\gray \scriptsize \begin{rotate}{90}{$\cdots$} \end{rotate}}
       \put(-3,23){\gray \scriptsize $1$}
       \put(31,12){ $\approx$}
       \put(46,12){ $\bE$}
       \put(62,12){ $\times$}
       \put(69,6.75){\gray \scriptsize $N$}
       \put(83,12){$\bA$}
       \put(69.5,19.5){\gray \scriptsize $1$}

       \put(71,12){\gray \scriptsize \begin{rotate}{90}{$\cdots$} \end{rotate}}
       \put(83,22){\gray \scriptsize $\cdots$}

       \put(0,27){\gray \scriptsize $1$}
       \put(11,27){\gray \scriptsize $\cdots$}
       \put(25,27){\gray \scriptsize $T$}
       \put(41,27){\gray \scriptsize $1$}
       \put(47,27){\gray \scriptsize $\cdots$}
       \put(56,27){\gray \scriptsize $N$}
       \put(72,22){\gray \scriptsize $1$}
       \put(97,22){\gray \scriptsize $T$}
  \end{overpic}
  \caption{The linear NMF model: $\bX \approx \bE \bA $, under the constraints $\bE \geq 0$ and $\bA \geq 0$. Throughout this paper, $t=1,2, \ldots, T$ and $n=1,2, \ldots, N$, where the factorization rank $N$ is assumed to be known or estimated using any off-shelf technique \cite{Kanagal10}.}
  \label{fig:NMF}
\end{figure}

\begin{figure}[t]
\centering
\scriptsize
  \begin{overpic}[width=0.49\textwidth,]{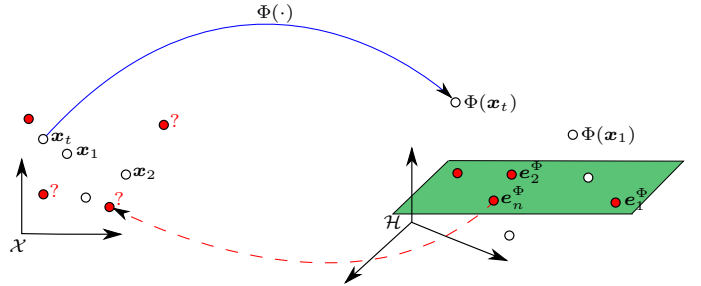}
     \put(9,19.5){$\bx_1$}
     \put(5.25,21.5){$\bx_t$}
     \put(17.5,16.25){$\bx_2$}
     \put(5,13.25){\red $?$}
     \put(14.75,12){\red $?$}
     \put(23,23.75){\red $?$}
     \put(67,26.5){$\Phi(\bx_{t})$}
     \put(84.5,21.75){$\Phi(\bx_{1})$}
     \put(91,12){$\be^\Phi_1$}
     \put(75.25,16){$\be^\Phi_2$}
     \put(72.5,12.25){$\be^\Phi_n$}
     \put(36,40){$\Phi(\cdot)$}
     \put(-1,5){$\cp{X}$}
     \put(55,8.5){$\cp{H}$}
  \end{overpic}
  \caption{Illustration of the straightforward application of the NMF in the feature space, as studied in \cite{Zhang2006,Ding10,Li2012}. All elements $\be^\Phi_n$, for $n=1,2, \ldots, N$, belong to the feature space $\cp{H}$ spanned by the images $\Phi(\bx_{t})$, for $t=1,2, \ldots, T$. One has no access to these elements, nor to their pre-images (shown with~{\red$?$}) in the input space $\cp{X}$.}
  \label{fig:NMF_old}
\end{figure}

In this paper, we propose an original kernel-based framework for nonlinear NMF that does not suffer from the curse of the pre-image problem, as opposed to other techniques derived within kernel machines (see \figurename~\ref{fig:NMF_old} and \figurename~\ref{fig:NMF_new} for a snapshot of this difference). To this end, we explore a novel model defined by the mapping of the columns of the matrices (the investigated matrix and the first unknown one), these columns lying in the input space. It turns out that the corresponding optimization problem can be efficiently tackled directly in the input space, thanks to the nature of the underlying kernel function. We derive two iterative algorithms: an additive update rule based on a gradient descent scheme, and a multiplicative update rule in the same spirit of \cite{lee99}. We investigate expressions associated to the polynomial and Gaussian kernels, as well as the linear one which yields the conventional linear NMF. Based on the proposed framework, we describe several extensions to incorporate constraints, including sparseness and smoothness, as well as a TV-like spatial regularization. The relevance of the proposed approach with its extensions is shown on two well-known hyperspectral images
. We also provide a theoretical analysis of the (non)convexity of the studied optimization problem, with connections to previous work such as \cite{polyNMF}, as given in the Appendix.

The rest of the paper is organized as follows: First, the NMF is presented in its ubiquitous form, demonstrating the difficulty of applying the NMF in the feature space. Section~\ref{sec:KNMF} describes the proposed framework for kernel-NMF. Several extensions of the kernel-NMF are developed in Section~\ref{sec:extensions} for incorporating constraints. Section~\ref{sec:experiments} illustrates the relevance of the proposed techniques for unmixing two real hyperspectral images Cuprite and Moffett. Section~\ref{sec:conclusion} concludes this paper with future work. 

\section{The NMF, from linear to kernel}\label{sec:NMF}

This section presents the conventional linear NMF and its kernel-based counterparts, illustrating the 
pre-image problem.

\subsection{A primer on the NMF}

The conventional NMF consists in approximating a nonnegative matrix $\bX$ with a product of two nonnegative matrices $\bE$ and $\bA$, namely
\begin{equation}\label{eq:NMF0}
	\bX \approx \bE \bA
\end{equation}
subject to $\bE \geq 0$ and $\bA \geq 0$; See Figure \ref{fig:NMF} for notations. The former nonnegativity constraint is relaxed in the so-called semi-NMF. The optimization problem is written in terms of the nonnegative least squares optimization, with $\arg\min_{\bA, \bE\geq 0} \frac12 \|\bX - \bE \bA\|^2_F$, where $\|\cdot\|_F$ is the Frobenius norm.

Under the nonnegativity constraints, the estimation of the entries of both matrices $\bE$ and $\bA$ is not convex. Luckily, the estimation of each matrix, separately, is a convex optimization problem. Most NMF algorithms take advantage of this property, with an iterative technique that alternates the optimization over each matrix while keeping the other one fixed. The most commonly used algorithms are the gradient descent rule and the multiplicative update rule (expressions are given in Section~\ref{sec:kernels.linear}). See \cite[Chapter 13]{ICAhandbook10} for a recent survey of several standard algorithms. See also \cite{NMF_WhyHow} and references therein.

It is easy to notice that the matrix model \eqref{eq:NMF0} can be considered vector-wise, by dealing separately with each column of the matrix $\bX$. Let $\bX = [ \bx_1 ~~ \bx_2 ~ \cdots ~ \bx_T ]$, $\bE = [\be_1 ~~ \be_2 ~ \cdots ~ \be_N]$, and $a_{nt}$ be the $(n,t)$-th entry in $\bA$. Then the NMF consists in estimating the nonnegative vectors $\be_n$ and scalars $a_{nt}$, for all $n=1, 2, \ldots, N$ and $t=1, 2, \ldots, T$, such that
\begin{equation}\label{eq:NMF}
    \bx_t \approx \sum_{n=1}^N a_{nt} \, \be_n.
\end{equation}
Following this model, the resulting optimization problem is $\arg\min_{a_{nt}, \be_n \geq 0} \frac12 \sum_{t=1}^T\|\bx_t - \sum_{n=1}^N a_{nt} \, \be_n\|^2$. It is this vector-wise model that is investigated in deriving kernel-based NMF.

Without loss of generality, we illustrate the NMF with the problem of unmixing in hyperspectral imagery. In this case, following the notation in \eqref{eq:NMF}\footnote{It is worth noting that the NMF model is symmetric, that is \eqref{eq:NMF0} is equivalent to $\bX^\top \approx \bA^\top \bE^\top$. In other words, the meaning of abundance matrix and endmember matrix is somewhat arbitrary in the definition \eqref{eq:NMF0}.}, each spectral $\bx_t$ of the image is decomposed into a set of spectra $\be_1, \be_2, \ldots, \be_N$ ({\em i.e.,} endmembers), while $a_{1t}, a_{2t}, \ldots, a_{Nt}$ denote their respective abundances. Such physical problem allows us to incorporate additional constraints and impose structural regularity of the solution, as detailed in Section~\ref{sec:extensions}.

\begin{figure}[t]
\centering
\scriptsize
  \begin{overpic}[width=0.49\textwidth,]{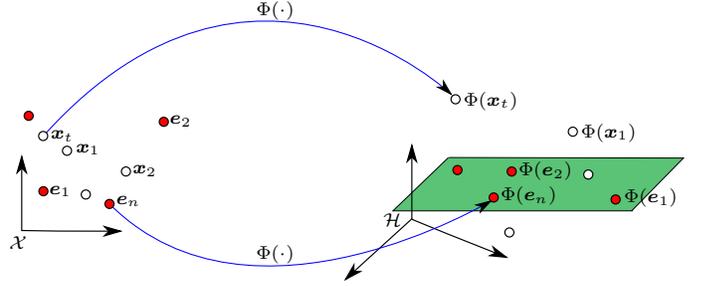}
     \put(9,19.5){$\bx_1$}
     \put(5.25,21.5){$\bx_t$}
     \put(17.5,16.25){$\bx_2$}
     \put(5,13.25){$\be_1$}
     \put(15,11.75){$\be_n$}
     \put(23,23.75){$\be_2$}
     \put(67,26.5){$\Phi(\bx_{t})$}
     \put(84.5,21.75){$\Phi(\bx_{1})$}
     \put(91,12){$\Phi(\be_1)$}
     \put(75.25,16){$\Phi(\be_2)$}
     \put(72.5,12.25){$\Phi(\be_n)$}
     \put(36,40){$\Phi(\cdot)$}
     \put(36,3.5){$\Phi(\cdot)$}
     \put(-1,5){$\cp{X}$}
     \put(55,8.5){$\cp{H}$}
  \end{overpic}
  \caption{Illustration of the kernel-NMF proposed in this paper. As opposed to the one shown in \figurename~\ref{fig:NMF_old}, the proposed approach estimates the elements $\be_n$ directly in $\cp{X}$, which is the input space. This strategy allows to overcome the curse of the pre-image problem, by estimating directly the spectra.}
  \label{fig:NMF_new}
\end{figure}

\subsection{On kernelizing the NMF: the pre-image problem}

Few attempts have been made to derive nonlinear, kernel-based, NMF. These methods originate in mapping the columns of $\bX$ with a nonlinear function $\Phi(\cdot)$, namely transforming $\bx_t$ into $\Phi(\bx_t)$ for  $t=1, 2, \ldots, T$. Let $\cp{H}$ be the resulting feature space, with the associated norm $\|\Phi(\bx_{t})\|_\cp{H}$ and the corresponding
inner product $\langle \Phi(\bx_{t}) , \Phi(\bx_{t'}) \rangle_\cp{H}$. The latter defines the so-called kernel function $\kappa(\bx_{t},\bx_{t'})$ in  kernel machines.

Written in the feature space, the NMF model is
\begin{equation}\label{eq:kNMFold}
	\Phi(\bx_t)  \approx \sum_{n=1}^N a_{nt} \, \be_n^\Phi,
\end{equation}
written in matrix form as $\bX^\Phi  \approx \big[\be_1^\Phi ~~ \be_2^\Phi ~ \cdots ~ \be_N^\Phi \big] \, \bA$, where $\bX^\Phi = \big[\Phi(\bx_1) ~~ \Phi(\bx_2) ~ \cdots ~ \Phi(\bx_T)\big]$. Here, the elements $\be_n^\Phi$ lie in the feature space $\cp{H}$, since $\Phi(\bx_t)$ belongs to the span of all $\be_n^\Phi$. Essentially, all kernel-based NMF proposed so far have been considering this model \cite{Zhang2006,polyNMF,Ding10,Li2012,an2011multiple}. Unfortunately, the model \eqref{eq:kNMFold} suffers from an important weakness, inherited from kernel machines: one has no access to the elements in the feature space, but only to their inner products with the kernel function. The fact that the elements $\be_n^\Phi$ lie in the feature space $\cp{H}$ leads to several drawbacks in NMF, as shown next.

Back to the model \eqref{eq:kNMFold}, one has for all $t,t'=1, 2, \ldots, T$:
\begin{equation*}
	\langle \Phi(\bx_{t'}),\Phi(\bx_{t}) \rangle_\cp{H}
		\approx \sum_{n=1}^N a_{nt} \, \langle \Phi(\bx_{t'}) , \be_n^\Phi \rangle_\cp{H}.
\end{equation*}
Here, the left-hand-side is equivalent to $\kappa(\bx_{t'},\bx_t)$. Unfortunately, the inner product $\langle \Phi(\bx_{t'}) , \be_n^\Phi \rangle_\cp{H}$ cannot be evaluated using the kernel function. To circumvent this difficulty, one should restrict the form of $\be_n^\Phi$, as investigated in \cite{Lee2009,an2011multiple} where the authors write them in terms of a linear combination of $\Phi(\bx_{t})$. By rearranging the coefficients of the linear combination in a matrix $\bW$, the problem takes the form $\bX^\Phi \approx \bX^\Phi \bW \bA$. While this simplifies the optimization problem, it is however quiet different from the conventional NMF problem \eqref{eq:NMF0}.

Another downside of the model \eqref{eq:kNMFold} is that one cannot impose the nonnegativity of the elements in the feature space, and in particular $\be_n^\Phi$. Therefore, the constraint $\be_n^\Phi \geq 0$ should be dropped. Only the coefficients $a_{nt}$ can be set to nonnegative values. In this case, one can no longer tackle the NMF problem, but the relaxed semi-NMF problem, where only the constraint $\bA \geq 0$ is imposed \cite{Li2012}.

The most important drawback is that one has no access to the elements $\be_n^\Phi$. Having a given matrix $\bX$, only the matrix $\bA$ is determined. To estimate a matrix $\bE$, one needs to solve the so-called pre-image problem. This ill-posed problem consists of estimating an input vector whose image, defined by the nonlinear map $\Phi(\cdot)$, is as close as possible to a given element in the feature space \cite{11.spm}. In other words, one determines each column $\be_n$ of $\bE$ by solving $\Phi(\be_n) \approx \be_n^\Phi$, for all $n=1, 2, \ldots, N$, which is a non-convex, non-linear, ill-posed problem. This issue is obvious in all previous work on kernel-based NMF; see for instance \cite{MercerNMF}. Including the nonnegativity constraint to the pre-image problem is a challenging problem, as investigated in our recent work \cite{KalEUSIPCO11,13.pr.nn_preimage}.

Few attempts were conducted to circumvent some of these difficulties. The homogeneous kernel is considered in \cite{polyNMF}, restricting the derivation to this kernel as argued by the authors; see Appendix for more details. The authors of \cite{MercerNMF} approximate the kernel by one associated to a {\em nonnegative map}, which requires to solve another optimization problem prior to processing the one associated to the NMF. Moreover, the pre-image problem needs to be solved subsequently.


For all these reasons, applying the nonnegative matrix factorization in the feature space has been often limited to preprocessing data before solving a classification problem. 
Still, one has no access to the bases in the resulting relevant representation. Next, we propose a  framework where both matrices can be exhibited, without suffering from the curse of the pre-image problem. The core of the difference between these two approaches is illustrated in \figurename~\ref{fig:NMF_old} and \figurename~\ref{fig:NMF_new}.

\section{A novel framework for kernel-NMF}\label{sec:KNMF}
\label{sec:KNMF}

In this section, we propose a novel framework to derive kernel-NMF, where the  underlying model is defined by entries in the input space, and therefore without the pain of solving the pre-image problem. To this end, we explore the characteristics of the investigated kernel.

We consider the following matrix factorization model:
\begin{equation*}
    \bX^{\Phi} \approx \bE^{\Phi} \bA.
\end{equation*}
where $\bE^\Phi = [\Phi(\be_1) ~~ \Phi(\be_2) ~ \cdots ~ \Phi(\be_N)]$. The nonnegativity constraint is imposed to $\bA \geq 0$ and $\be_n \geq 0$ for all $n=1, 2, \ldots, N$. One can also consider the semi-NMF variant. Therefore, we have the following model:
\begin{equation}\label{eq:KNMF}
    \Phi(\bx_t) \approx \sum_{n=1}^N a_{nt} \, \Phi(\be_n).
\end{equation}
This means that we are estimating the elements $\be_n$ directly in the input space, as opposed to the model given in \eqref{eq:kNMFold} where the elements $\be_n^\Phi$ lie in the feature space.

To estimate all $\be_n$ and $a_{nt}$, we consider a simple alternating technique to minimize the cost function
\begin{equation}\label{eq:prob}
   J=
   \frac{1}{2} \sum_{t=1}^T \Big\| \Phi(\bx_t) - \sum_{n=1}^N a_{nt} \, \Phi(\be_n) \Big\|_{\cp{H}}^2,
\end{equation}
thus yielding the optimization problem
\begin{equation*}
    \min_{a_{nt}, \be_n} \sum_{t=1}^T \Big( - \sum_{n=1}^N a_{nt} \kappa(\be_n,\bx_t) + \frac{1}{2} \sum_{n=1}^N \sum_{m=1}^N a_{nt} a_{mt} \kappa(\be_n, \be_{m}) \Big),
\end{equation*}
where $\kappa(\bx_t, \bx_t)$ is removed from the expression since it is independent of $a_{nt}$ and $\be_n$. By taking its derivative  with respect to $a_{nt}$, we obtain the following expression:
\begin{equation*}
    \nabla\!_{a_{nt}} J = - \kappa(\be_n,\bx_t) + \sum_{m=1}^N a_{mt} \, \kappa(\be_n, \be_{m}).
\end{equation*}
By taking the gradient of $J$ with respect to $\be_n$, we obtain:
\begin{equation}\label{eq:grad_e}
    \nabla\!_{\be_n} J = \sum_{t=1}^T a_{nt} \Big ( - \nabla\!_{\be_n} \kappa(\be_n,\bx_t) + \sum_{m=1}^N a_{mt} \, \nabla\!_{\be_n} \kappa(\be_n, \be_{m}) \Big ).
\end{equation}
Here, $\nabla\!_{\be_n} \kappa(\be_n,\cdot)$, which denotes the gradient of the kernel with respect to its argument $\be_n$, can be easily derived for most valid kernels, as given in \cite{KalEUSIPCO11,13.pr.nn_preimage} for a problem different from the NMF. See Section~\ref{sec:kernels} for the case of the linear, polynomial and Gaussian kernels. But before, we derive two iterative algorithms for solving the above kernel-NMF, by alternating the estimation of $a_{nt}$ and $\be_n$.

\subsection{Algorithms}

\subsubsection{Additive update rule}

In the first iterative algorithm, an additive update rule is presented to solve the optimization problem. It is based on a gradient descent scheme, alternating over both $a_{nt}$ and $\be_n$, and is followed by a rectification function to impose their nonnegativity. A normalization step to impose sum-to-one constraint on $\ba_t$ can also be used.

By using a gradient descent scheme, we update $a_{nt}$ according to
    $a_{nt} = a_{nt} - \eta_{nt} \, \nabla\!_{a_{nt}} J$,
where the stepsize $\eta_{nt}$ can take different values for each pair $(n,t)$. Replacing $\nabla\!_{a_{nt}} J$ with its expression, we get the following update rule:
\begin{equation}\label{eq:updates}
    a_{nt} = a_{nt} - \eta_{nt} \Big( \sum_{m=1}^N a_{mt} \, \kappa(\be_n, \be_{m}) - \kappa(\be_n,\bx_t) \Big).
\end{equation}
A similar procedure is applied to estimate the elements $\be_n$. The obtained update rule is given by
\begin{equation}\label{eq:updatea}
    \be_{n} = \be_{n} - \eta_{n} \nabla\!_{\be_{n}}J ,
\end{equation}
where the stepsize $\eta_{n}$ can depend on $n$, and the expression of $\nabla\!_{\be_{n}} J$ is given in \eqref{eq:grad_e}. To impose the nonnegativity of the matrices, the negative values obtained by the above update are set to zero. This is done by using the rectification function $x = \max (x, 0)$ over all $a_{nt}$ and the entries in all the vectors $\be_n$. The abundance vectors can be normalized to have unit $\ell_1$-norm, by substituting $\ba_{t}$ with ${\ba_{t}}/{\|\ba_t\|_1}$.

\subsubsection{Multiplicative update rule}

The additive update rule is a simple procedure, however, the convergence is generally slow, and is directly related to the stepsize value used. In order to overcome these issues, we propose a multiplicative update rule, in the same spirit as in the conventional NMF \cite{Lee2001}.

To derive a multiplicative update rule for $a_{nt}$, the stepsize $\eta_{nt}$ in \eqref{eq:updates} is chosen such that the first and the third terms in its right-hand-side cancel, that is
\begin{equation*}
	\eta_{nt} = \frac{a_{nt}}{ \sum_{m=1}^N a_{mt} \, \kappa(\be_n, \be_{m})}.
\end{equation*}
Therefore, by substituting this expression in \eqref{eq:updates}, we get the following update rule:
\begin{equation}\label{eq:updatesmul}
   a_{nt} = a_{nt} \times \frac{ \kappa(\be_n,\bx_t)}{ \sum_{m=1}^N a_{mt} \, \kappa(\be_n, \be_{m}) }.
\end{equation}
A normalization $\ba_{t}={\ba_{t}}/{\|\ba_t\|_1}$ can be considered to satisfy the sum-to-one constraint. Compared with the additive rule, the above multiplicative rule has several interesting properties, such as the absence of any tunable stepsize parameter and the nonexistence of any rectification function. The latter property is due to the
multiplicative nature which ensures that elements cannot become negative when one initializes with a nonnegative right-hand-side of \eqref{eq:updatesmul}.

A similar procedure is applied to estimate the elements $\be_n$, for $n=1, 2, \ldots,N$. The trick is that the expression of the gradient \eqref{eq:grad_e} can always be decomposed as $\nabla\!_{\be_n} J = P - Q$, where $P$ and $Q$ have nonnegative entries. This is called the split gradient method \cite{Lanteri11}. It is obvious that this decomposition is not unique. Still, one can provide a multiplicative update for a given kernel function, as shown next.


\subsection{Kernels}\label{sec:kernels}

All kernels studied in the literature about kernel machines can be investigated in our framework. In the following, we derive expressions of the update rules for the most known kernel functions.

\subsubsection{Back to the conventional linear NMF}\label{sec:kernels.linear}

A key property of the proposed kernel-NMF framework is that the conventional NMF is a special case, when the linear kernel is used with $\kappa(\be_n, \bz) = \bz^\top \be_n$, for any vector $\bz$ from the input space. The gradient of the kernel is $\nabla\!_{\be_n} \kappa(\be_n, \bz) = \bz$ in this case. By substituting this result in the above expressions, we get the additive update rules
\begin{equation*}
	\left\{\begin{array}{r@{\;}l}
	a_{nt} =&  a_{nt} - \eta_{nt} \Big(\sum_{m=1}^N a_{mt} \, \be_m^\top \be_{n} - \bx_t^\top \be_n \Big);\\
    {\be_{n}} =&  {\be_{n}} - \eta_{n} \sum_{t=1}^T a_{nt} \Big ( - \bx_t + \sum_{m=1}^N a_{mt} \, \be_{m} \Big ),
	\end{array}\right.
\end{equation*}
as well as the multiplicative update rules
\begin{equation}\label{eq:mult_Linear}
	\left\{\begin{array}{l@{}l}
	a_{nt} &=  \displaystyle a_{nt} \times \frac{\bx_t^\top \be_n}{ \sum_{m=1}^N a_{mt} \, \be_m^\top \be_{n} };\\
    {\be_{n}} &=  \displaystyle {\be_{n}} \otimes \frac{\sum_{t=1}^T a_{nt} \, \bx_t}{{\sum_{t=1}^T} a_{nt} \sum_{m=1}^N a_{mt} \, \be_{m}}.
	\end{array}\right.
\end{equation}
In the latter expression for updating $\be_{n}$, the element-wise operations are used, with the division and multiplication, the latter being the Hadamard product given by $\otimes$. These expressions yield the well-known classical NMF. It is worth noting that in the case of the linear kernel, namely when the map $\Phi(\cdot)$ is the identity operator, the optimization problem \eqref{eq:prob} is equivalent to the minimization of the (half) Frobenius norm between the matrices $\bX$ and $\bE \bA$.


\subsubsection{The polynomial kernel}

The polynomial kernel is defined as $\kappa(\be_n, \bz) = (\bz^\top \be_n+ c)^d$. Here, $c$ is a nonnegative constant balancing the impact of high-order to low-order terms in the kernel. The kernel's gradient is given by:
\begin{equation*}
\nabla\!_{\be_n}\kappa(\be_n, \bz) = d \, (\bz^\top \be_n+ c)^{(d-1)}\bz.
\end{equation*}
We consider the most common quadratic polynomial kernel with $d=2$.
Replacing $\nabla\!_{\be_n}\kappa(\be_n, \bz)$ with this result, we obtain the additive update rules
\begin{equation*}
	\left\{\begin{array}{r@{\;}l}
	a_{nt}   =&  a_{nt} - \eta_{nt} \Big(\sum_{m=1}^N a_{mt} (\be_m^\top \be_{n}+c)^2 - (\bx_t^\top\be_n + c)^2\Big);\\
    {\be_{n}} =&  {\be_{n}} - \eta_{n} \sum_{t=1}^T a_{nt} \Big( -2 (\bx_t^\top\be_n  + c) \, \bx_t \\
    		&\qquad\qquad\qquad\qquad\:\, + 2\sum_{m=1}^N a_{mt} (\be_m^\top\be_n + c) \, \be_m\Big ),
	\end{array}\right.
\end{equation*}
and the multiplicative update rules
\begin{equation}\label{eq:mult_Poly}
	\left\{\begin{array}{l@{}l}
	a_{nt} &=  \displaystyle a_{nt} \times \frac{(\bx_t^\top\be_n +c)^2}{ \sum_{m=1}^N a_{mt} \, (\be_m^\top \be_n+c)^2 };\\
    {\be_{n}} &=  \displaystyle {\be_{n}} \otimes \frac{\sum_{t=1}^T a_{nt} (\bx_t^\top\be_n +c)\bx_t}{{\sum_{t=1}^T} a_{nt} \sum_{m=1}^N a_{mt} (\be_m^\top \be_n+c) \be_{m}}.
	\end{array}\right.
\end{equation}


\subsubsection{The Gaussian kernel}

The Gaussian kernel is defined by $\kappa(\be_n, \bz) = \exp (\frac{-1}{2 \sigma^2} \| \be_n - \bz \; \|^2)$. In this case, its gradient is
\begin{equation*}
\nabla\!_{\be_n} \kappa(\be_n, \bz) = -\frac{1}{\sigma^2} \kappa(\be_n,\bz) (\be_n - \bz).
\end{equation*}
The update rules of $a_{nt}$ can be easily derived, in both additive and multiplicative cases. For the estimation of $\be_{n}$, the additive rule is
\begin{align*}
    {\be_{n}} =  \be_{n} - \eta_{n} \Big( &
    + \frac{1}{\sigma^2}\sum_{t=1}^T a_{nt} \, \kappa(\be_n,\bx_t) (\be_n - \bx_t)\\
     &
      - \frac{1}{\sigma^2}\sum_{t=1}^T \sum_{m=1}^N a_{nt} a_{mt} \, \kappa(\be_n, \be_{m}) (\be_n - \be_m) \Big ).
\end{align*}
As for the multiplicative algorithm, we split the corresponding gradient into the subtraction of two terms with nonnegative entries. This is possible since all the matrices are nonnegative, as well as the kernel values. We get the update rule:
\begin{equation}\label{eq:mult_Gauss}
    \be_{n} \!=\! \displaystyle \be_{n} \otimes \frac{ \sum_{t=1}^T  a_{nt} \Big (\bx_t \kappa(\be_n,\bx_t)  + \sum_{m=1}^N a_{mt} \be_n \kappa(\be_n, \be_{m}) \Big )} {{\sum_{t=1}^T} a_{nt} \Big (\be_n \kappa(\be_n,\bx_t) + \sum_{m=1}^N a_{mt} \be_{m} \kappa(\be_n, \be_{m}) \Big )},
\end{equation}
where the division is component-wise. 

\section{Extensions of kernel-NMF}\label{sec:extensions}

The above work provides a framework to derive extensions of the kernel-NMF by including some constraints and incorporating structural information. Several extensions are described in the following with constraints imposed on the endmembers and the abundances, typically motivated by the unmixing problem in hyperspectral imagery defined by the model in \eqref{eq:KNMF}. 

\subsection{Constraints on the endmembers}

Different constraints can be imposed on the endmembers, essentially to improve the smoothness of the estimates. It turns out that the derivatives, with respect to the abundances, of the unconstrained cost function $J$ in \eqref{eq:prob} and the upcoming constrained cost functions are identical. Thus, the resulting update rules for the estimation of the abundances remain unchanged, as detailed in \eqref{eq:updates} for the additive scheme and \eqref{eq:updatesmul} for the multiplicative scheme.

\subsubsection{Smoothness with 2-norm regularization}

In the estimation of $\be_n$, one is interested in regular solutions, namely with less variations, {\em e.g.}, less spiky \cite{Piper2011}. This property is exploited by the so-called smoothness constraint, by minimizing $\frac12 \sum_{n=1}^N \|\be_{n}\|^{2}$ in the input space. By combining this penalty term with the cost function \eqref{eq:prob}, we get
\begin{equation*}
    J_\textrm{2-norm}= \frac{1}{2} \sum_{t=1}^T \| \Phi(\bx_t) - \sum_{n=1}^N a_{nt} \, \Phi(\be_n) \|_{\cp{H}}^2+\frac{\lambda}{2}\sum_{n=1}^N \|\be_{n}\|^{2}.
\end{equation*}
The parameter $\lambda$ controls the balance between the reconstruction accuracy (first term in the above expression) and the smoothness of all $\be_{n}$ (second term).

To estimate the endmember $\be_n$, we consider the gradient of $J_\textrm{2-norm}$ with respect to it, which yields the following additive update rule:
\begin{align*}
   {\be_n} \!=\!  \be_n  \!- \eta_{n}\Big(&\sum_{t=1}^T  a_{nt}
   \Big(\sum_{m=1}^N  a_{mt}\nabla\!_{\be_n}\kappa(\be_n, \be_{m}) -\! \nabla\!_{\be_n} \kappa(\be_n,\bx_t)\Big)
   \\ &+\lambda{\be_n}\Big).
\end{align*}
Using the split gradient method \cite{Lanteri11}, we get the corresponding multiplicative update rule. It turns out that one gets the same expressions as in the unconstrained case, with \eqref{eq:mult_Linear}, \eqref{eq:mult_Poly} or \eqref{eq:mult_Gauss}, where the term $\lambda \, \be_n$ is added to the denominator.


We can also consider a similar constraint in the feature space within the kernel-NMF framework. The cost function becomes
\begin{equation*}
    J_\textrm{2-norm}^{\cp{H}}= \frac{1}{2} \sum_{t=1}^T \| \Phi(\bx_t) - \sum_{n=1}^N a_{nt} \, \Phi(\be_n) \|_{\cp{H}}^2+\frac{\lambda_{\cp{H}}}{2}\sum_{n=1}^N \|\be_{n}\|_{\cp{H}}^{2}.
\end{equation*}
The gradient with respect to $\be_n$ yields the additive update rule
\begin{align*}
   {\be_n} \!=\! {\be_n}  \!- \eta_{n}\Big(&\sum_{t=1}^T a_{nt} \Big(\sum_{m=1}^N a_{mt}\nabla\!_{\be_n}\kappa(\be_n, \be_{m}) -\! \nabla\!_{\be_n} \kappa(\be_n,\bx_t)\Big) \\
   & +\lambda_{\cp{H}}\nabla\!_{\be_n} \kappa(\be_n, \be_n) \Big).
\end{align*}
Depending on the used kernel, the expression of the multiplicative update rule is similar to the one given in the unconstrained case, with \eqref{eq:mult_Linear}, \eqref{eq:mult_Poly} or \eqref{eq:mult_Gauss}, by adding the term $\lambda_{\cp{H}} \nabla\!_{\be_n} \kappa(\be_n, \be_n)$ to the denominator.

%

It is easy to see that, when dealing with the linear kernel where $\nabla\!_{\be_n} \kappa(\be_n, \be_n) = \be_n$, the corresponding update rules are equivalent to the ones given with the constraint in the input space. Moreover, it turns out that smoothing in the feature space associated to the Gaussian kernel makes no sense, since 
$\nabla\!_{\be_n} \kappa(\be_n, \be_n)=0$.

\subsubsection{Smoothness with fluctuation regularization}

In \cite{virtanen2003sound}, Virtanen imposes smoothness on every endmember, in a sense that the fluctuations between neighboring values within $\be_i$ is small. The cost function of the kernel-NMF with such constraint is
\begin{equation*}
    {J_\textrm{fluct}} 
    \!=\! \frac{1}{2} \sum_{t=1}^T \| \Phi(\bx_t) - \sum_{n=1}^N a_{nt} \, \Phi(\be_n) \|_{\cp{H}}^2 \notag
    + \frac{\gamma}{2}\sum_{n=1}^N \sum_{l=2}^{L-1}|e_{ln}-e_{(l-1)n}|,
\end{equation*}
where $\gamma$ is a tradeoff parameter. The derivative of the penalizing term with respect to $e_{ln}$ equals to:
$$ \left\{
\begin{array}{rcl}
+ \gamma     & & \text{when } {{e_{ln}}<{e_{(l-1)n}} \textrm{ and } {e_{ln}}<{e_{(l+1)n}}} ;\\
- \gamma     & &  \text{when } {{e_{ln}}>{e_{(l-1)n}} \textrm{ and } {e_{ln}}>{e_{(l+1)n}}} ;\\
0            & & {\text{otherwise}}.
\end{array} \right. $$

Adopting the descent gradient scheme \eqref{eq:updatea} and incorporating the above expression into $\nabla\!_{\be_{n}} J $ given in \eqref{eq:grad_e}, we can easily get the modified additive and multiplicative update rules for the endmembers estimation. The corresponding expressions are omitted due to space limitation. 

\subsubsection{Smoothness with weighted-average regularization}\label{sec:smooth_averaged}

Another smoothness regularization raised by Chen and Cichocki in \cite{Cichocki} aims to reduce the difference between $e_{ln}$ and a weighted average $\overline{e}_{ln}=\alpha \overline{e}_{(l-1)n}+(1-\alpha){e_{ln}}$. For each $\be_{n}$, this can be written in a matrix form
as
  $\overline{\be}_{n} = \textbf{T} \, \be_{n}$,
where
\begin{equation*}
 {\textbf{T}} =
 \left(\begin{array}{cccc} 
 (1-\alpha)  &  0  &  \cdots  &  0  \\
 {\alpha \, (1-\alpha)} & (1-\alpha)  & {\cdots} & 0  \\
 \vdots &  & \ddots  & {\vdots} \\
 {\alpha}^{L-1}(1-\alpha) & \cdots &  {\alpha \, (1-\alpha)} & (1-\alpha)
 \end{array}\right).
\end{equation*}
For each ${\be_n}$, the cost function is defined as:
 ${\frac{1}{L}\|\be_{n}-\overline{\be}_{n}\|^2} = \frac{1}{L}\|(\textbf{I}-\textbf{T})\be_n\|^2$.
By considering all endmembers, for $n=1,2, \ldots, N$, and introducing a regularization parameter $\rho$ that controls the smoothing process, we get the cost function:
\begin{equation*}
   {J_\textrm{av}} 
            =  \frac{1}{2} \sum_{t=1}^T \| \Phi(\bx_t) - \sum_{n=1}^N a_{nt} \, \Phi(\be_n) \|_{\cp{H}}^2
         +\frac{\rho}{2L}\sum_{n=1}^N \|(\textbf{I}-\textbf{T}) \be_n\|^2.
\end{equation*}

The gradient of the penalty term with respect of $\be_n$ takes the form $\rho \, \textbf{Q} \be_{n}$, where $\textbf{Q }= \frac{1}{L}{(\textbf{I}-\textbf{T})}^\top(\textbf{I}-\textbf{T})$. The additive update rule of the endmembers is easy to derive using the descent gradient method. The multiplicative update rule depends on the used kernel, with expressions similar to \eqref{eq:mult_Linear}, \eqref{eq:mult_Poly} and \eqref{eq:mult_Gauss}, by adding the term $\rho \, \textbf{Q} \be_{n}$ to the denominator.


\subsection{Constraints on the abundances}

To satisfy a physical interpretation, two types of constraints are often imposed on the abundances, the sparseness and the spatial regularity. It turns out that the these constraints have no influence on the update rules for the endmembers estimation as given in Section~\ref{sec:KNMF}. As a consequence, we shall study in detail the estimation of the abundances.

\subsubsection{Sparseness regularization}

Sparseness has been proved to be very attractive in many disciplines, namely by penalizing the $\ell_1$-norm of the weight coefficients \cite{Hoyer04}. Typically in the hyperspectral unmixing problem, each spectrum $\bx_t$ can be represented by using a few endmembers, namely only a few abundances $a_{nt}$ are non-zero. Since the latter are nonnegative, the $\ell_1$-norm of their corresponding vector is $\sum_{n=1}^N a_{nt}$. This leads to the following sparsity-promoting cost function
\begin{equation*}
    J_\textrm{sparse}= \frac{1}{2} \sum_{t=1}^T \Big\| \Phi(\bx_t) - \sum_{n=1}^N a_{nt} \, \Phi(\be_n) \Big\|_{\cp{H}}^2+\mu \sum_{t=1}^T \sum_{n=1}^N a_{nt},
\end{equation*}
where the parameter $\mu$ controls the tradeoff between the reconstruction accuracy and the sparseness level. By considering the derivative of $J_\textrm{sparse}$ with respect to $a_{nt}$, the additive update rule is obtained as follows:
\begin{equation*}
  a_{nt}=a_{nt} - \eta_{nt} \big(\sum_{m=1}^N a_{mt} \, \kappa(\be_n, \be_{m}) - \kappa(\be_n,\bx_t)+ \mu \big).
\end{equation*}
To get the multiplicative update rule, we set the stepsize to $\eta_{nt} = a_{nt}/(\sum_{m=1}^N a_{mt} \,\kappa(\be_n, \be_m)+\mu)$, which leads to
\begin{equation*}\label{Sparseness_Multiplicative_Rule}
  a_{nt} = \displaystyle a_{nt} \times \frac{ \kappa(\be_n,\bx_t)}{\sum_{m=1}^N a_{mt} \, \kappa(\be_n, \be_{m}) + \mu }.
\end{equation*}

\subsubsection{Spatial regularization}

Spatial regularization that favors spatial coherence is essential in many image processing techniques, as often considered in the literature with the total-variation (TV) penalty. This penalty was recently studied in \cite{Iordache2012} for the linear unmixing problem in hyperspectral imagery. Motivated by this work, we derive in the following a TV-like penalty for incorporating spatial regularity within the proposed framework. It is worth noting that the derivations of the spatial regularization can be viewed as the application on the abundances of the method given in Section~\ref{sec:smooth_averaged}, by extending the one-direction smoothness (of $e_{ln}$) into the two-dimensional spatial regularization (of $a_{nk}$).

\begin{figure}[t]
\graphicspath{{Graphics/}}
\centering
\tiny
\psfrag{C}{$\cb{M}\!_{n}(i,j)$}
\psfrag{R}{$\!\!\cb{M}\!_{n}(i,j+1)$}
\psfrag{L}{$\!\cb{M}\!_{n}(i,j-1)$}
\psfrag{U}{$\!\!\cb{M}\!_{n}(i-1,j)$}
\psfrag{D}{$\!\!\cb{M}\!_{n}(i+1,j)$}
 \graphicspath{{Graphics/}}
\includegraphics[scale=0.35]{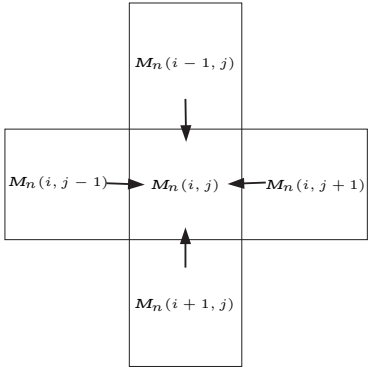}
\caption{Schematic illustration of the spatial regularization. ${\cb{M}\!_{n}(i,j)}$ represents the abundance of the {\emph{n}-th} endmember for the $(i,j)$-th pixel. Each of the four neighbors imposes a spatial regularization effect on the center pixel.}
\label{fig:spatial}
\end{figure}

When transforming ({\em i.e.}, folding) a hyperspectral image of size $T=a \times b$ pixels into a matrix $\bX$, the $t$-th column of $\bX$ is filled with the $(i,j)$-th spectrum from the original image, with $i=\lceil \frac{t}{b}\rceil$ and $j=t-(i-1)b$. In the following, we denote by $\cb{M}\!_n$ the matrix of the $n$-th abundance defined by the entries $\cb{M}\!_n(i,j)=a_{nk}$, with $k=(i-1)b+j$ for $i=1,2,\ldots,a$ and $j=1,2,\ldots,b$.
For any inner element $\cb{M}\!_n(i,j)$ belonging to the $n$-th abundance map, we shall use for spatial regularization the four geographical neighboring directions; cf. \figurename~\ref{fig:spatial}.

The four spatial weighted averages of ${\cb{M}\!_{n}}(i,j)$ from its left, right, up and down sides are denoted as ${\overline{\cb{M}\!_{n}}(i,j)}_{\rightarrow}$, ${\overline{\cb{M}\!_{n}}(i,j)}_{\leftarrow}$, ${\overline{\cb{M}\!_{n}}(i,j)}_{\downarrow}$ and ${\overline{\cb{M}\!_{n}}(i,j)}_{\uparrow}$. They are expressed as follows:
\begin{equation*}
\left\{
\begin{aligned}
&{\overline{\cb{M}\!_{n}}(i,j)}_{\rightarrow} = \alpha {\overline{\cb{M}\!_{n}}{(i,j-1)}_{\rightarrow} } + (1-\alpha) \cb{M}\!_{n}(i,j) \\
&{\overline{\cb{M}\!_{n}}(i,j)}_{\leftarrow} = \alpha {\overline{\cb{M}\!_{n}}{(i,j+1)}_{\leftarrow}} + (1-\alpha) \cb{M}\!_{n}(i,j) \\
&{\overline{\cb{M}\!_{n}}(i,j)}_{\downarrow} = \alpha {\overline{\cb{M}\!_{n}}{(i-1,j)}_{\downarrow}} + (1-\alpha) \cb{M}\!_{n}(i,j) \\
&{\overline{\cb{M}\!_{n}}(i,j)}_{\uparrow} = \alpha {\overline{\cb{M}\!_{n}}{(i+1,j)}_{\uparrow}} + (1-\alpha) \cb{M}\!_{n}(i,j). \\
\end{aligned}
\right.
\end{equation*}
Rewriting in matrix form, we get
\begin{equation*}
\left\{
\begin{aligned}
&{\overline{\cb{M}}_{n}^\top}(i,:)_{\rightarrow} ={\cb{T}\!_{\rightarrow}{\cb{M}^\top_{n}(i,:)}} \\
&{\overline{\cb{M}}_{n}^\top}(i,:)_{\leftarrow} ={\cb{T}\!_{\leftarrow}{\cb{M}^\top_{n}(i,:)}} \\
&{\overline{\cb{M}}_{n}}(:,j)_{\downarrow} = \cb{T}\!_{\downarrow}{\cb{M}\!_{n}(:,j)} \\
&{\overline{\cb{M}}_{n}}(:,j)_{\uparrow}  = \cb{T}\!_{\uparrow}{\cb{M}\!_{n}(:,j)},
\end{aligned}
\right.
\end{equation*}
where $\cb{T}\!_{\leftarrow} = {\textbf{T}^{\top}_{\rightarrow}}$, $\cb{T}\!_{\uparrow} = {\textbf{T}^{\top}_{\downarrow}}$, with
\begin{equation*}
 \cb{T}\!_{\rightarrow} =
 \left(\begin{array}{cccc} 
 (1-\alpha)  &  0  &  \cdots  &  0  \\
 {\alpha(1-\alpha)} & (1-\alpha)  & {\cdots} & 0  \\
 \vdots & \ddots & \ddots  & {\vdots} \\
 {\alpha}^{b-1}(1-\alpha) & \cdots &  {\alpha (1-\alpha)} & (1-\alpha)
 \end{array}\right),
\end{equation*}
\begin{equation*}
  \cb{T}\!_{\downarrow} =
 \left(\begin{array}{cccc} 
 (1-\alpha)  &  0  &  \cdots  &  0  \\
 {\alpha(1-\alpha)} & (1-\alpha)  & {\cdots} & 0  \\
 \vdots & \ddots & \ddots  & {\vdots} \\
 {\alpha}^{a-1}(1-\alpha) & \cdots &  {\alpha (1-\alpha)} & (1-\alpha)
 \end{array}\right).
\end{equation*}
For each abundance $\ba_n$, the associated cost function is:
\begin{align*}
    R_n \!\!=\! \tfrac{1}{2} \! \sum_{i=1}^a\sum_{j=1}^b
    &\frac{\omega_{l}}{b} \|(\textbf{I} \!-\! \cb{T}\!_{\rightarrow}){\cb{M}^\top_{n}}(i,:)\|^2
             \!\!+\! \frac{\omega_{r}}{b} \|(\textbf{I} \!-\! \cb{T}\!_{\leftarrow}){\cb{M}^\top_{n}}(i,:)\|^2
         \\
     \!\!+& \frac{\omega_{u}}{a} \|(\textbf{I} \!-\! \cb{T}\!_{\downarrow}){\cb{M}\!_{n}}(:,j)\|^2
           \!\!+\! \frac{\omega_{d}}{a} \|(\textbf{I} \!-\! \cb{T}\!_{\uparrow}){\cb{M}\!_{n}}(:,j)\|^2.
\end{align*}
Here, $\omega_{l},\omega_{r},\omega_{u}$ and $\omega_{d}$ control spatial effect ratios of left, right, up and down direction. In particular, $\omega_{l}=\omega_{r}=\omega_{u}=\omega_{d}$ denotes an average allocation of spatial effects. Considering the regularization term $\sum_{n=1}^N R_n$ for all $N$ abundance maps, the cost function of the spatially-regularized kernel-NMF is:
\begin{align}\label{eq:ContinuityS}
    {J_\textrm{spatial}}
            = & \frac{1}{2} \sum_{t=1}^T \| \Phi(\bx_t) - \sum_{n=1}^N a_{nt} \, \Phi(\be_n) \|^2
          + \sum_{n=1}^N{R_n}.
\end{align}
The update rule of the abundances for this cost function is obtained 
by locating $a_{nt}$ in $\cb{M}\!_n$ using $a_{nt}= \cb{M}\!_{n}(i,j)$, with $i=\lceil\frac{t}{b}\rceil$ and $j=t-(i-1)b$. We get
    $\textstyle {\nabla_{a_{nt}} (\sum_{n=1}^N R_n)} =
     {\nabla_{\cb{M}\!_n(i,j)}{R_n}}
           = \cb{G}(i,j)$,
where
\begin{equation*}
{\cb{G}}=\omega_{l}\cb{M}\!_{n}\textbf{Q}_{\rightarrow}
	+\omega_{r}\cb{M}\!_{n}\textbf{Q}_{\leftarrow}
	+\omega_{u}\cb{M}\!_{n}^\top\textbf{Q}_{\downarrow}
	+\omega_{d}\cb{M}\!_{n}^\top\textbf{Q}_{\uparrow}
\end{equation*}
with
\begin{equation*}
\left\{
\begin{aligned}
&{\textbf{Q}_{\rightarrow}} = {\tfrac{1}{b}{(\textbf{I}-\cb{T}\!_{\rightarrow})}^\top(\textbf{I}-\cb{T}\!_{\rightarrow})}   \\
&{\textbf{Q}_{\leftarrow}} = {\tfrac{1}{b}{(\textbf{I}-\cb{T}\!_{\leftarrow})}^\top(\textbf{I}-\cb{T}\!_{\leftarrow})}    \\
&{\textbf{Q}_{\downarrow}} = {\tfrac{1}{a}{(\textbf{I}-\cb{T}\!_{\downarrow})}^\top(\textbf{I}-\cb{T}\!_{\downarrow})}    \\
&{\textbf{Q}_{\uparrow}} = {\tfrac{1}{a}{(\textbf{I}-\cb{T}\!_{\uparrow})}^\top(\textbf{I}-\cb{T}\!_{\uparrow})}.
\end{aligned}
\right.
\end{equation*}
By computing ${\nabla_{a_{nt}}{J_\textrm{spatial}}}$ with the above expression, we get the additive update rule for $a_{nt}$:
\begin{equation*}\label{eq:updateA}
 a_{nt} = a_{nt} - \eta_{nt} \big(\sum_{m=1}^N a_{mt} \, \kappa(\be_n, \be_{m}) - \kappa(\be_n,\bx_t)+ \cb{G}(i,j) \big),
\end{equation*}
as well as the multiplicative update rule, where we use ${\eta_{nt}} = {a_{nt}}/\big({\sum_{m=1}^{N} a_{mt}\kappa(\be_n,\be_m)+\cb{G}(i,j)}\big)$:
\begin{equation*}\label{eq:updateA}
 a_{nt} = a_{nt} \times \frac{\kappa(\be_n,\bx_t)}{ \sum_{m=1}^N a_{mt} \, \kappa(\be_n, \be_{m})+ \cb{G}(i,j) }.
\end{equation*}

\section{Experiments}\label{sec:experiments}

In this section, the relevance of the derived kernel-NMF and its extensions is studied on real hyperspectral images. The studied images are well-known \cite{HaAlDoTo2011}, acquired by the Airborne Visible/Infrared Imaging Spectrometer (AVIRIS). The raw images consists of 244 spectral bands, with the wavelength ranging from $0.4\mu m$ to $2.5\mu m$. The first image is a sub-image of $50\times50$ pixels taken from the well-known Cuprite image, where $L=189$ spectral bands (out of 244) are of interest. The geographic composition of this area is known to be dominated by muscovite, alunite and cuprite, as investigated in \cite{clark1993mapping}. The second image is from the Moffett Field image, with a studied sub-image of $50\times50$ pixels. This scene is known to consist of three materials: vegetation, soil and water. Before analysis, the noisy and water absorption bands were removed, yielding $L=186$ spectral bands as recommended in \cite{Dobigeon2008}.

We introduce two criteria to evaluate the unmixing performance. Reconstruction error in the input space (RE) measures the mean distance between any spectrum and its reconstruction using the estimated endmembers and abundances, with
\begin{equation*}
   \text{RE}=\sqrt{\frac{1}{TL}\sum_{t=1}^T \|\bx_{t}-{\sum_{n=1}^{N}a_{nt}\be_{t}}\|^2}. \notag
\end{equation*}
Similarly, the reconstruction error in the feature space is
\begin{equation*}
     \text{RE}^\Phi=\sqrt{\frac{1}{TL}\sum_{t=1}^T \|\Phi (\bx_{t})-{\sum_{n=1}^{N}a_{nt}\Phi(\be_{t})}\|_\cp{H}^2}.
\end{equation*}

\subsection{State-of-the-art methods}

Most state-of-the-art unmixing algorithms either extract the endmembers (such as with VCA and N-Findr) or estimate the abundances (such as with FCLS, and nonlinear K-Hype and GBM-sNMF). In this case, the solving the unmixing problem requires the join use of two algorithms, one for endmember extraction and one for abundance estimation. The proposed kernel-NMF estimates simultaneously the endmembers and the abundances, in the same spirit as some recently developed algorithms (such as MinDisCo and ConvexNMF). In the following, we succinctly present all the comparing algorithms.


The endmember extraction is often operated separately of the abundance estimation. The commonly used techniques are the N-Findr \cite{N-FINDR} and the vertex component analysis (VCA) \cite{VCA}. These techniques rely on the linear unmixing model and assume the existence of endmembers in the image. They are convex-geometry-based techniques that inflate the simplex formed by the spectra, where the endmembers correspond to the vertices of the largest simplex englobing the spectra. Since they provide comparable results, they are used whenever needed by the abundance estimation techniques.

The most known abundance estimation technique is the fully constrained least squares algorithm (FCLS) \cite{Heinz}. By considering the linear mixing model, it is a least square technique that estimates the abundances under the nonnegativity and sum-to-one constraints. Nonlinear unmixing with the estimation of the abundances has been recently investigated, with a model that has two terms, a conventional linear mixing model and an additive nonlinear one. In \cite{13.tsp.unmix}, the nonlinearity is defined using a kernel-based formulation, yielding the linear-mixture/nonlinear-fluctuation model (K-Hype). More recently, the generalized bilinear model is considered in \cite{Yokoya14}, and solved using a semi-nonnegative matrix factorization (GBM-sNMF). All these techniques require a complete knowledge of the endmembers, identified with either N-Findr or VCA.

We also considered two non-kernel techniques that jointly extract the endmembers and estimate the abundances. The minimum dispersion constrained NMF (MinDisCo) \cite{Huck} integrates the dispersion regularity into the NMF, by minimizing the variance of each endmember and imposing the sum of abundance fractions for every pixel to converge to 1. The resulting problem is solved with an alternate projected gradient scheme. In terms of convex optimization, the convex NMF (ConvexNMF) proposed in \cite{Ding10} restricts the basic matrix (endmember matrix in our problem) by nonnegative linear combinations of samples, thus facilitating the interpretation.

Furthermore, we compared to other kernel-based NMF approaches. Kernel convex-NMF (KconvexNMF) and kernel semi-NMF based on nonnegative least squares (KsNMF), are the kernelized methods corresponding respectively to the ConvexNMF in \cite{Ding10} and the alternating nonnegativity constrained least squares with the active set method in \cite{Kim2008}, as proposed in \cite{Li2012}. 
Due to the curse of the pre-image in the methods studied in \cite{Zhang2006,Li2012}, neither the endmembers can be represented explicitly nor the reconstruction error can be evaluated. As opposed to these methods, the Mercer-based NMF introduced in \cite{MercerNMF} (MercerNMF) provides comparable results. It is based on constructing a Mercer kernel that has a kernel map close to the one from the Gaussian kernel, under the nonnegative constraint on the embedded data. Conventional NMF is finally performed on these mapped data. It is noteworthy that learning the nonnegative embedding is computationally expensive.


\subsection{Search for the appropriate parameters}

To provide comparable results, we estimated the optimal values of the parameters by conducting experiments on the unconstrained kernel-NMF with the multiplicative scheme (denoted by Poly$\otimes$ and Gauss$\otimes$), since the latter does not depend on the stepsize parameter as in the case of the additive scheme (denoted by Poly$\oplus$ and Gauss$\oplus$). In order to explore the influence brought by the different regularizations to the unmixing performance, we used the same parameter values in the case of the constrained extensions of the kernel-NMF. Note that the number of iterations was set to 200 for all experiments.

\begin{figure}[t]
\centering
\subfigure[Cuprite image]{
\label{Fig.FindSigmaCuprite}
 \graphicspath{{Graphics/}}
 \psfragscanon
 \psfrag{RE}{\footnotesize RE}
 \psfrag{f}{\scriptsize~~$^\Phi$}
 \psfrag{s}{\scriptsize$\sigma$}
 \includegraphics[trim = 25mm 5mm 15mm 0mm, clip, width=.45\textwidth]{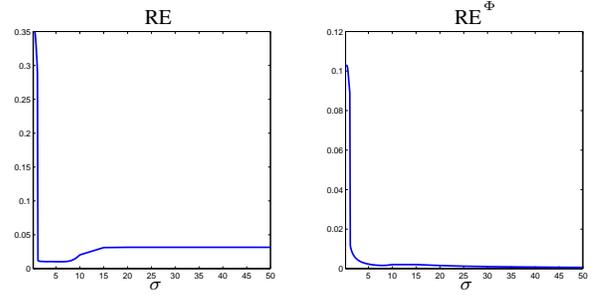}}
\subfigure[Moffett image]{
\label{Fig.FindSigmaMoffett}
 \graphicspath{{Graphics/}}
 \psfragscanon
 \psfrag{RE}{\footnotesize RE}
 \psfrag{f}{\scriptsize~~$^\Phi$}
 \psfrag{s}{\scriptsize$\sigma$}
 \includegraphics[trim = 25mm 5mm 15mm 0mm, clip, width=.45\textwidth]{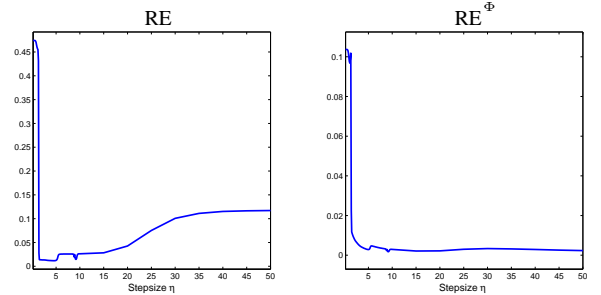}}
\caption{Influence on the reconstruction errors of the Gaussian bandwidth parameter $\sigma$ for the unconstrained kernel-NMF with the multiplicative update rules.}
\label{Fig.FindSigma}
\end{figure}

In the case of the polynomial kernel, we used the quadratic kernel with $d=2$ since it is related to the generalized bilinear model as suggested in \cite{13.tsp.unmix}. The influence of the additive constant $c$ is illustrated in \figurename~\ref{Fig.FindConstant}, yielding $c=0.44$ for the Cuprite and $c=0.72$ for the Moffett scene. 
A similar process was taken to determine the bandwidth parameter $\sigma$ of the Gaussian kernel, employing the same candidate values set $\{0.2, 0.3, \ldots, 9.9, 10, 15, 20, \ldots, 50$\} for both images. The reconstruction errors are shown in \figurename~\ref{Fig.FindSigma}. We fixed $\sigma=2.5$ and $\sigma=3.3$ for the Cuprite and the Moffett images, respectively. 

Concerning the stepsize parameter in the additive scheme, it is not only image-wise, but also involves a tradeoff between the estimation accuracy and the convergence rate.

\begin{figure}[t]
\label{Fig.FindConstant}
\centering
\subfigure[Cuprite image]{
\label{Fig.FindConstantCuprite}
 \graphicspath{{Graphics/}}
 \psfragscanon
 \psfrag{RE}{\footnotesize RE}
 \psfrag{f}{\scriptsize~~$^\Phi$}
 \psfrag{c }{\scriptsize $c$}
 \psfrag{c}{\scriptsize $c$}
\includegraphics[trim = 25mm 5mm 15mm 0mm, clip, width=.45\textwidth]{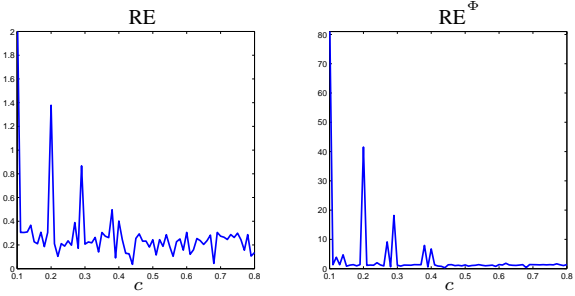}}
\subfigure[Moffett image]{
\label{Fig.FindConstantMoffett}
 \graphicspath{{Graphics/}}
 \psfragscanon
 \psfrag{RE}{\footnotesize RE}
 \psfrag{f}{\scriptsize~~$^\Phi$}
 \psfrag{c }{\scriptsize $c$}
 \psfrag{c}{\scriptsize $c$}
 \includegraphics[trim = 25mm 5mm 15mm 0mm, clip, width=.45\textwidth]{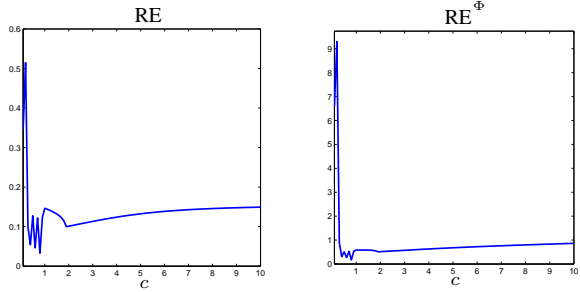}}
\caption{Influence on the reconstruction errors of the parameter $c$ of the polynomial kernel for the unconstrained kernel-NMF with the multiplicative update rules.}\label{Fig.FindConstant}
\end{figure}

\begin{table}[t]
\renewcommand{\arraystretch}{1.4}
\scriptsize
\centering 
\caption{Unmixing performance}\label{Tab.BasicNMFMoffett}
\begin{tabular}{@{}|@{\;}c@{\,}|l|c|c|c|c|}
\cline{3-6}
\multicolumn{2}{c|}{}&\multicolumn{2}{c|}{Cuprite}&\multicolumn{2}{c|}{Moffett} \\
\cline{3-6}
\multicolumn{2}{c|}{}&RE \tiny$\times {10}^{-2}$\! & $\text{RE}^\Phi$\tiny$\times {10}^{-2}$\! 			& RE \tiny$\times {10}^{-2}$\! & $\text{RE}^\Phi$\tiny$\times {10}^{-2}$\!   \\
\cline{2-6}
\multicolumn{1}{c|}{}&{FCLS}&3.20&-&15.61&-\\
\cline{2-6}
\multicolumn{1}{c|}{}&{K-Hype}&2.12&-&5.27&-\\
\cline{2-6}
\multicolumn{1}{c|}{}&{GBM-sNMF}&0.98&-&2.09&-\\
\cline{2-6}
\multicolumn{1}{c|}{}&{MinDisCo}&1.65&-&2.92&-\\
\cline{2-6}
\multicolumn{1}{c|}{}&{ConvexNMF}&1.61&-&2.58&\\
\cline{2-6}
\multicolumn{1}{c|}{}&{KconvexNMF}&-&25.64&-&35.95\\
\cline{2-6}
\multicolumn{1}{c|}{}&{KsNMF}&-&1.38&-&2.30\\
\cline{2-6}
\multicolumn{1}{c|}{}&{MercerNMF}&-&2.74&-&2.77\\

\hline
\multirow{6}{*}{\rotatebox{90}{ this paper\;}} & {Lin$\oplus$}&0.96&0.96&2.90&2.90\\
\cline{2-6}
& {Lin$\otimes$} & \bf 0.93 &0.93&\bf 0.73 &0.73\\
\cline{2-6}
& Poly$\oplus$&5.61&31.80&7.53&33.52\\
\cline{2-6}
& Poly$\otimes$&3.60&30.59&2.68&14.85\\
\cline{2-6}
&Gauss$\oplus$&2.16&0.94&2.12&0.98\\
\cline{2-6}
&Gauss$\otimes$ & 1.05 & {\bf 0.50} & 1.24 & {\bf 0.45}\\
\hline
\end{tabular}
\end{table}

\begin{figure}[t]
\graphicspath{{Graphics/Basic/Cuprite/}}
\begin{minipage}{.98\linewidth}
\centerline{(a) Linear kernel }
\includegraphics[trim = 35mm 35mm 30mm 30mm, clip,width=.95\textwidth]{End_LM.eps}
\\
\includegraphics[trim = 35mm 35mm 30mm 35mm, clip,width=1\textwidth]{Ab_LM.eps}
\centerline{(b) Polynomial kernel }
\includegraphics[trim = 35mm 35mm 30mm 30mm, clip,width=.95\textwidth]{End_PM.eps}
\\
\includegraphics[trim = 35mm 35mm 30mm 35mm, clip,width=1\textwidth]{Ab_PM.eps}
\centerline{(c) Gaussian kernel }
\includegraphics[trim = 35mm 35mm 30mm 30mm, clip,width=.95\textwidth]{End_GM.eps}
\\
\includegraphics[trim = 35mm 35mm 30mm 35mm, clip,width=1\textwidth]{Ab_GM.eps}
\end {minipage}
\caption{Cuprite image: Endmembers and corresponding abundance maps, estimated by the unconstrained kernel-NMF with Lin$\otimes$, Poly$\otimes$ and Gauss$\otimes$ update rules.}
\label{Fig.EndAbunBasicCuprite}
\bigskip
\end{figure}

\begin{figure}[t]
\graphicspath{{Graphics/Basic/Moffett/}}
\smallskip
\begin{minipage}{.98\linewidth}
\centerline{(a) Linear kernel }
\includegraphics[trim = 35mm 35mm 30mm 30mm, clip,width=.95\textwidth]{End_LM.eps}
\\
\includegraphics[trim = 35mm 35mm 30mm 35mm, clip,width=1\textwidth]{Ab_LM.eps}
\smallskip
\centerline{(b) Polynomial kernel }
\includegraphics[trim = 35mm 35mm 30mm 30mm, clip,width=.95\textwidth]{End_PM.eps}
\\
\includegraphics[trim = 35mm 35mm 30mm 35mm, clip,width=1\textwidth]{Ab_PM.eps}
\centerline{(c) Gaussian kernel }
\includegraphics[trim = 35mm 35mm 30mm 30mm, clip,width=.95\textwidth]{End_GM.eps}
\\
\includegraphics[trim = 35mm 35mm 30mm 35mm, clip,width=1\textwidth]{Ab_GM.eps}
\end {minipage}
\medskip
\caption{Moffett image: Endmembers and corresponding abundance maps, estimated by the unconstrained kernel-NMF with Lin$\otimes$, Poly$\otimes$ and Gauss$\otimes$ updates rules.}
\label{Fig.EndAbunBasicMoffett}
\bigskip
\end{figure}

\subsection{Performance of the 
kernel-NMF}

Experiments were conducted on the linear (Lin$\oplus$/Lin$\otimes$), the polynomial (Poly$\oplus$/Poly$\otimes$) and the Gaussian (Gauss$\oplus$/Gauss$\otimes$) kernels. The endmembers and the corresponding abundance maps estimated using these algorithms are shown in \figurename~\ref{Fig.EndAbunBasicCuprite} for the Cuprite image and in \figurename~\ref{Fig.EndAbunBasicMoffett} for the Moffett image. The efficiency of the kernel-NMF is compared to the aforementioned well-known unmixing techniques, as presented in \tablename~\ref{Tab.BasicNMFMoffett}.

Despite the fact that the linear kernel leaded to small reconstruction error in the input space, it does not outperform the Gaussian kernel in the feature space. As reflected in \figurename~\ref{Fig.EndAbunBasicCuprite}, the inherent nonlinear correlation of the Cuprite image is revealed using the Gaussian kernel, which recognizes the three regions in the abundance maps; whereas linear kernel is only capable to distinguish two regions. Considering the reconstruction error in the feature space, the unconstrained kernel-NMF with the Gaussian kernel surpasses not only its counterparts with the linear and the polynomial kernels, but also all other methods including the kernel-based ones.

\begin{figure}[t]
\graphicspath{{Graphics/FluctuationRegularization/}}
\includegraphics[trim = 20mm 20mm 20mm 20mm, clip,width=.5\textwidth]{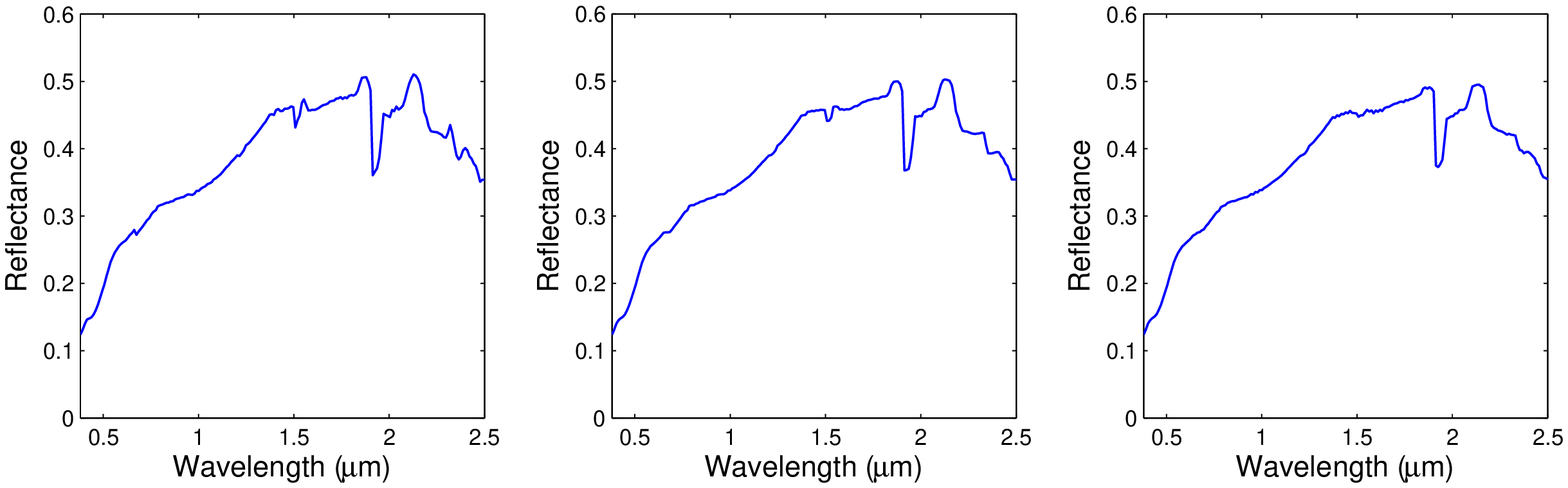}
~\\[-3.4cm]
{\tiny\mbox{}\qquad\qquad\,$\boxed{\gamma = 0}$
\qquad\qquad\qquad\qquad\qquad\quad~\;$\boxed{\!\gamma = 0.1\!}$
\qquad\qquad\qquad\qquad\qquad~~~$\boxed{\!\gamma = 0.2\!}$
}\\[-.07cm]
{\red \mbox{}\qquad\qquad~\;\,$\bigcirc$
\qquad\qquad\qquad\quad$\bigcirc$
\qquad\qquad\qquad~~~$\bigcirc$
}\\[-.25cm]
{\red \mbox{}\qquad\qquad\qquad\quad\;$\bigcirc$
\qquad\qquad\qquad\quad$\bigcirc$
\qquad\qquad\qquad\quad$\bigcirc$\!\!\!
}
\\[1.5cm]
%
\caption{Influence of the smoothness with fluctuation regularization, illustrated on an endmember estimated from the Cuprite image, with different values of the regularization parameter $\gamma$.}\label{Fig.fluctCuprite}
\bigskip
\end{figure}
\begin{figure}[t]
\graphicspath{{Graphics/WeightedAverageRegularization/}}
\centerline{\includegraphics[trim = 20mm 20mm 20mm 20mm, clip,width=.5\textwidth]{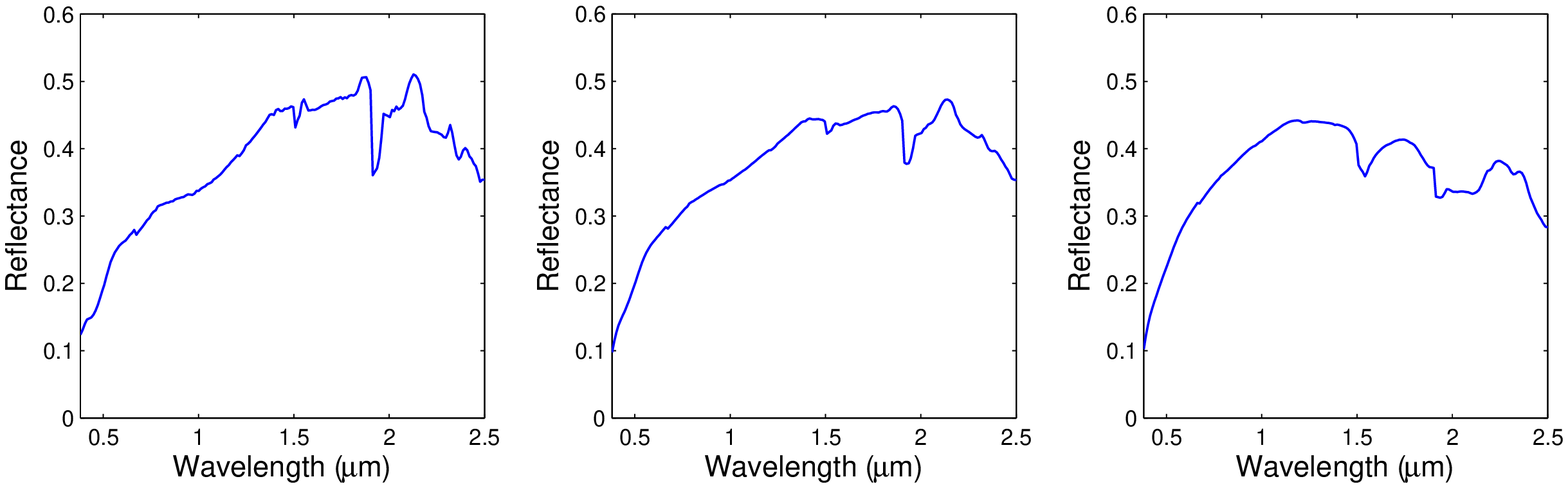}}
~\\[-1.8cm]
{\tiny\mbox{}\qquad\qquad\qquad\quad\,$\boxed{~\rho = 0~}$
\qquad\qquad\qquad\qquad\qquad\,$\boxed{~\rho = 4000~}$
\qquad\qquad\qquad\qquad\quad$\boxed{~\rho = 6000~}$
}
\\[0.425cm]
\caption{Influence of the weighted-average regularization, illustrated on an endmember estimated from the Cuprite image, with different values of the regularization parameter $\rho$.}\label{Fig.averagedCuprite}
\bigskip\bigskip
\end{figure}
\begin{figure*}[t]
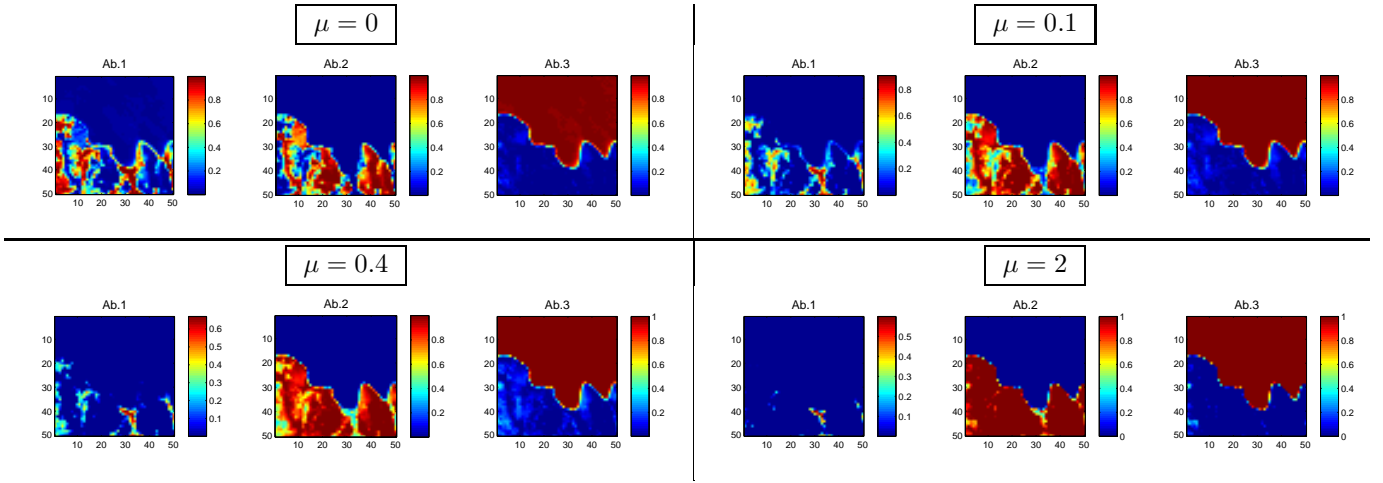

\graphicspath{{Graphics/SparsenessRegularization/Moffett/}}
\begin{minipage}{0.5\linewidth}
\centerline{$\boxed{~\mu=0~}$}
\centerline{\includegraphics[trim = 20mm 30mm 20mm 30mm, clip,width=1\textwidth]{Regu0.eps}}
\end{minipage}
\hfill \vrule
\begin{minipage}{0.5\linewidth}
\centerline{$\boxed{~\mu=0.1~}$}
\centerline{\includegraphics[trim = 20mm 30mm 20mm 30mm, clip,width=1\textwidth]{Regu0p1.eps}}
\end{minipage}
\vfill \hrule
\begin{minipage}{0.5\linewidth}
\centerline{}
\centerline{$\boxed{~\mu=0.4~}$}
\centerline{\includegraphics[trim = 20mm 30mm 20mm 30mm, clip,width=1\textwidth]{Regu0p4.eps}}
\end{minipage}
\hfill \vrule
\begin{minipage}{0.5\linewidth}
\centerline{}
\centerline{$\boxed{~\mu=2~}$}
\centerline{\includegraphics[trim = 20mm 30mm 20mm 30mm, clip,width=1\textwidth]{Regu2.eps}}
\end{minipage}
\caption{Influence of the sparseness regularization of the abundance maps for the Moffett image.}\label{Fig.sparseMoffett}
\bigskip
\end{figure*}
\begin{figure*}
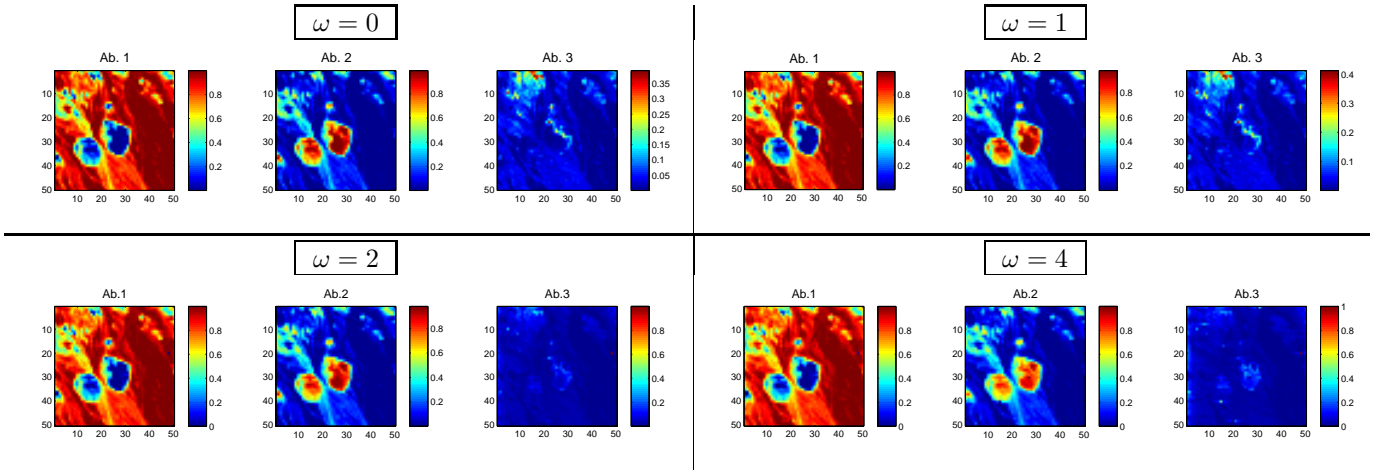

\graphicspath{{Graphics/SpatialRegulazation/Cuprite/}}
\begin{minipage}{0.5\linewidth}
\centerline{$\boxed{~\omega=0~}$}
\centerline{\includegraphics[trim = 20mm 30mm 20mm 30mm, clip,width=1\textwidth]{Regu0.eps}}
\end{minipage}
\hfill \vrule
\begin{minipage}{0.5\linewidth}
\centerline{$\boxed{~\omega=1~}$}
\centerline{\includegraphics[trim = 20mm 30mm 20mm 30mm, clip,width=1\textwidth]{Regu1.eps}}
\end{minipage}
\vfill \hrule
\begin{minipage}{0.5\linewidth}
\centerline{}
\centerline{$\boxed{~\omega=2~}$}
\centerline{\includegraphics[trim = 20mm 30mm 20mm 30mm, clip,width=1\textwidth]{Regu2.eps}}
\end{minipage}
\hfill \vrule
\begin{minipage}{0.5\linewidth}
\centerline{}
\centerline{$\boxed{~\omega=4~}$}
\centerline{\includegraphics[trim = 20mm 30mm 20mm 30mm, clip,width=1\textwidth]{Regu4.eps}}
\end{minipage}
\caption{Influence of the spatial regularization of the abundance maps for the Cuprite image, with $\alpha=0.5$.}\label{Fig.spatialCuprite}
\bigskip
\end{figure*}

We also conducted an analysis on the different extensions. The results corresponding to the proposed regularizations are detailed in \figurename~\ref{Fig.fluctCuprite} and \figurename~\ref{Fig.averagedCuprite} for the smoothness of the endmembers, while constraints on the abundance maps are shown in \figurename~\ref{Fig.sparseMoffett} for the sparseness regularization and \figurename~\ref{Fig.spatialCuprite} for the spatial regularization.

\section{Conclusion}\label{sec:conclusion}

In this paper, we presented a new kernel-based NMF, where the matrices are estimated in the input space. By exploring the nature of the used kernel functions, this approach circumvents the curse of the pre-image problem. Additive and multiplicative update rules were proposed, and several extensions were derived in order to incorporate constraints such as sparseness, smoothness and spatial regularity. The efficiency of these techniques was illustrated on well-known real hyperspectral images. As for future work, we are extending this approach for dimensionality reduction such as the principal component analysis. Other kernel functions are investigated, as well as the choice of the parameters.

%



\appendix[On the (non)convexity of the kernel-NMF]
\label{sec:appendix}

In the conventional NMF, the estimation of each matrix, separately, is a convex suboptimization problem. The convergence of the multiplicative update (MU) rules for the linear NMF was studied in~\cite{Lee2001, Lin07}, where the monotone decreasing property is proved by constructing an auxiliary function as upper bound.
Similarly, the convergence of the MU rules was studied for the convex and semi-NMF \cite{Ding10} and projective NMF \cite{YangPNMF}.

Being practically efficient with relevant results, the MU rules have been extensively considered in the literature, including several nonlinear and kernel-based formulations of the NMF. Unfortunately, a convergence analysis cannot be investigated with auxiliary functions\footnote{Early work in \cite{polyNMF} restricts to the case of the polynomial kernel with $c=0$, and considers using the auxiliary function approach by limiting the range of input data $\bx$ and basis $\be$ (endmembers in our context). However, the proof in its \cite[APPENDIX \uppercase\expandafter{\romannumeral2}]{polyNMF} for basis update relies on the inequality $(\bx^\top\be)^{d-2} \geq (\be^{\top}\be)^{d-2}$, $\forall x\in [0, 255], e \in [0, 1]$. Unfortunately, this relation does not hold in general. Counterexamples include the cases where $\bx$ and $\be$ are orthogonal to each other.
}. While the proposed framework overcomes the pre-image problem (which is inherently nonconvex), MU rules still do not guarantee a monotone decrease of the cost function \eqref{eq:prob}, due to its nonconvexity in terms of $\be_n$ for an arbitrary nonlinear kernel. Indeed, we show in the following that the corresponding suboptimization problem is possibly nonconvex, for polynomial and Gaussian kernels, by proving that the corresponding Hessian matrix is no longer guaranteed to be positive semidefinite.

Before proceeding, we recall the following well-established results in the literature. Let $J\in \cp{C}^2$ be a function with continuous partial derivatives of first and second order on a convex set $\cp{S}$, and $\bH(\bx)$ its Hessian evaluated at $\bx$. 
\begin{definition}
$\bH(\bx)$ is called positive semidefinite on $\cp{S}$ if it is symmetric and satisfies $\bx^{\top} \bH\bx \geq 0 ~ \forall \bx\in \cp{S}$.
\end{definition}

\begin{proposition}
The function $J$ is convex if and only if $\bH(\bx)$ is positive semidefinite for all $\bx \in \cp{S}$.
\end{proposition}

\begin{proposition}
If a matrix $\bH$ is positive semidefinite then all its diagonal entries are nonnegative.
\end{proposition}

Let $\bH(\be_n)$ be the Hessian of the cost function \eqref{eq:prob} at $\be_n$. To show the nonconvexity, we determine some $\be_n$, such that at least one negative entry exists on the diagonal of $\bH(\be_n)$.

When the polynomial kernel $\kappa(\be_n, \bz) = (\bz^\top \be_n+ c)^d$ is used, the corresponding cost function \eqref{eq:prob} is possibly nonconvex.
To show this, we write
the $k$-th diagonal entry of $\bH(\be_n)$, namely
\begin{align*}
H_{kk} =
\frac{\partial J(\be_n)}{\partial e_{kn}^2}
&=d(d-1)\sum_{t=1}^T a_{nt}\Big(-(\bx_t^\top \be_n+ c)^{(d-2)} x_{kt}^2
\\&~~+\sum_{j=1}^N a_{jt} (\be_j^\top \be_n+ c)^{(d-2)} e_{jk}^2 \Big).
\end{align*}
One can easily find examples that yield negative values to this expression. For example, consider $d=2$. In this case, if $\sum_{t=1}^T a_{nt}(-x_{kt}^2+\sum_{j\neq n} a_{jt} e_{jk}^2)<0$, the problem is nonconvex since there exist some $\be_n$ that yields $H_{kk}(\be_n)<0$ ({\em e.g.}, set $e_{nk}$ small enough to make $H_{kk}(\be_n)$ negative).

These results on the polynomial kernel can be extended to the Gaussian, where the $k$-th diagonal entry of $\bH(\be_n)$ is
\begin{align*}
 {H}_{kk}=& 
 \frac{1}{\sigma^4}\sum_{t=1}^T\big(\sigma^2\kappa(\be_n,\bx_t)-\sigma^2\sum_{j=1}^N a_{jt} \kappa(\be_n,\be_j)\\
&+\sum_{j=1}^N a_{jt} \kappa(\be_n,\be_j)(e_{kn}-e_{kj})^2 - \kappa(\be_n,\bx_t)(e_{kn}-x_{kt})^2 \big).
\end{align*}
Examples that yield the nonconvex property can be easily found.
\section*{Acknowledgment}
This work was supported by the French ANR, grant HYPANEMA: ANR-12-BS03-0003.

\bibliographystyle{IEEEtran}
\bibliography{bib_fei,bibdesk_Paul,biblio_ph}

\end{document}